\documentclass{article}

\usepackage{hyperref}
\usepackage{caption}
\usepackage{url}
\usepackage{pifont}
\usepackage{multirow}
\usepackage[most]{tcolorbox}
\usepackage{enumitem}
\usepackage{booktabs}
\usepackage{wrapfig}
\usepackage{makecell}

\usepackage{pifont}
\newcommand{\cmark}{\ding{51}}%
\newcommand{\xmark}{$\times$}%

\usepackage{adjustbox} 
\usepackage{subcaption} 

\usepackage{graphicx}

\usepackage{microtype}
\usepackage{graphicx}
\usepackage{subcaption}
\usepackage{booktabs} 

\usepackage{hyperref}

\usepackage[preprint]{icml2026}

\usepackage{amsmath}
\usepackage{amssymb}
\usepackage{mathtools}
\usepackage{amsthm}

\usepackage[capitalize,noabbrev]{cleveref}

\theoremstyle{plain}

\theoremstyle{definition}

\theoremstyle{remark}

\usepackage[textsize=tiny]{todonotes}

\icmltitlerunning{ Imagine a City: CityGenAgent for 
  Procedural 3D City Generation}

\begin{document}

\twocolumn[
  \icmltitle{   Imagine a City: CityGenAgent for 
  Procedural 3D City Generation}



  \icmlsetsymbol{equal}{*}

  \begin{icmlauthorlist}
    \icmlauthor{Zishan Liu}{lingnan,comp}
    \icmlauthor{Zecong Tang}{comp,zju}
    \icmlauthor{Ruocheng Wu}{comp} 
    \icmlauthor{Xinzhe Zheng}{lingnan} \\
    \icmlauthor{Jingyu Hu}{huawei}
    \icmlauthor{Ka-Hei Hui}{autodesk}
    \icmlauthor{Haoran Xie}{lingnan}
    \icmlauthor{Bo Dai}{comp,hku}
    \icmlauthor{Zhengzhe Liu}{lingnan}
  \end{icmlauthorlist}

  \icmlaffiliation{lingnan}{Lingnan University, Hong Kong SAR, China}
    \icmlaffiliation{zju}{Zhejiang University, Hangzhou, China}
  \icmlaffiliation{comp}{Feeling AI, Shanghai, China}
    \icmlaffiliation{huawei}{The Chinese University of Hong Kong, Hong Kong SAR,  China}
     \icmlaffiliation{autodesk}{Autodesk AI Lab, Canada}
    \icmlaffiliation{hku}{The University of Hong Kong, Hong Kong SAR, China}

  \icmlcorrespondingauthor{Bo Dai}{bdai@hku.hk}

  \icmlkeywords{Machine Learning, ICML}

  \vskip 0.3in
]



\printAffiliationsAndNotice{Work partially done during an internship at Feeling AI.}

\begin{abstract}
The automated generation of interactive 3D cities is a critical challenge with broad applications in autonomous driving, virtual reality, and embodied intelligence. While recent advances in generative models and procedural techniques have improved the realism of city generation, existing methods often struggle with high-fidelity asset creation, controllability, and manipulation.
In this work, we introduce CityGenAgent, a natural language-driven framework for hierarchical procedural generation of high-quality 3D cities. Our approach decomposes city generation into two interpretable components, \textbf{Block Program} and \textbf{Building Program}. To ensure structural correctness and semantic alignment, we adopt a two-stage learning strategy:
(1) Supervised Fine-Tuning (SFT). We train BlockGen and BuildingGen to generate valid programs that adhere to schema constraints, including non-self-intersecting polygons and complete fields;
(2) Reinforcement Learning (RL). We design Spatial Alignment Reward to enhance spatial reasoning ability and Visual Consistency Reward to bridge the gap between textual descriptions and the visual modality. Benefiting from the programs and the models' generalization, CityGenAgent supports natural language editing and manipulation. Comprehensive evaluations demonstrate superior semantic alignment, visual quality, and controllability compared to existing methods, establishing a robust foundation for scalable 3D city generation. Demos are available at \href{https://citygenagent.github.io/}{our project page}.
\end{abstract}

\section{Introduction}
Interactive world models~\citep{team2025hunyuanworld} have become a prominent research direction, facilitating notable progress in 3D scene generation. These models have found broad applications in robotics simulation~\citep{wang2024grutopia, ren2025simworld}, game asset development~\citep{hu2024scenecraft, maleki2024procedural}, and virtual reality~\citep{10.1007/978-3-319-39907-2_69, ocal2024sceneteller, wen20253d}. However, generating city-scale scenes is particularly challenging due to the complexity of road networks, the diversity of building structures, and the presence of numerous urban facilities.

Procedural generation, which refers to the automatic creation of content through algorithmic processes, has a long history in video games and computer graphics. Traditional approaches~\citep{10.1145/383259.383292_cityenigne, inproceedings, benevs2014procedural} employ rule-based systems to generate road networks and buildings but they require considerable manual intervention and  significant labor expenses. The recent breakthroughs in deep learning have driven substantial progress in methods based on implicit representations and neural rendering~\citep{10.5555/3600270.3601875_SGAM, lin2023infinicity, xie2024generative, xie2024citydreamer}, enabling the synthesis of photorealistic imagery for city-scale environments. Nevertheless, these approaches still struggle to produce consistent and precise 3D geometry, which constrains their practical deployment in downstream simulation tasks and limits their flexibility for controllable scene editing.

\begin{table*}[t]
\fontsize{8pt}{10pt}\selectfont
\caption{Summary and Comparison of  3D City Generation.}
   \vspace{-2ex}
\label{summary_city}
\begin{center}
\begin{tabular}{cc|cccc}
%
\hline
\textbf{Type} & \textbf{Method} &\textbf{Text Input}& \textbf{Native 3D Output} & \textbf{Hierarchical Decomposition}  & \textbf{Manipulation}\\

\hline
\multirow{2}{*}{Rendering-based} & 
InifiCity~\citep{lin2023infinicity}       & \xmark  & NeRF  & \xmark & \xmark\\
& CityDreamer~\citep{xie2024citydreamer}        & \xmark & NeRF & \xmark & \xmark\\
\hline
\multirow{2}{*}{Diffusion-based} & CityGen~\citep{deng2025citygen}       & \xmark  & NeRF  & \xmark & \xmark\\
& WonderJourney~\citep{yu2024wonderjourney}       & \cmark & Point Cloud & \xmark& \xmark\\
\hline
\multirow{4}{*}{Procedure-based} & 3D-GPT~\citep{sun20253d}       & \cmark & Mesh & \xmark & \xmark\\
& CityCraft~\citep{deng2024citycraft}            & \cmark & Mesh & \xmark & \xmark \\
& UrbanWorld~\citep{shang2024urbanworld}      & \cmark & Mesh & \xmark & \xmark\\
& CityGenAgent(Ours)   & \cmark & Mesh & \cmark & \cmark\\
\hline
\end{tabular}
\end{center}
    \vspace{-5ex}
\end{table*}

Some studies have explored combining the linguistic priors and reasoning capabilities of Large Language Models (LLMs) with procedural generation techniques to enhance the output quality and structural richness of generated environments. Such efforts have been applied to both natural scenes~\citep{duan2025latticeworld, sun20253d} and indoor scenes~\citep{feng2023layoutgpt, yang2024holodeck}. In the context of city scene generation, CityCraft~\citep{deng2024citycraft}, UrbanWorld~\citep{shang2024urbanworld}, and CityX~\citep{zhang2024cityx} illustrate the potential of integrating LLMs into procedural content generation workflows to produce more coherent, scalable, and controllable urban environments. These methods mainly directly prompt LLMs and rely on retrieving fixed assets for placement rather than enabling LLMs to perform spatial reasoning and understanding in generation tasks. This limitation hampers models' ability to reliably adhere to input conditions and reduces flexibility for iterative editing or creative modifications. In addition, while some indoor datasets~\citep{fu20213d, zhang2025m3dlayout, zhong2025internscenes} exist to support embodied intelligence research, large-scale outdoor datasets, especially for city-scale scenes, are still lacking. The large scale and complexity of urban environments make data acquisition difficult, which further complicates the generation of high-quality and controllable city scenes. Therefore, the design of compact representations and dedicated generative models for city-scale scene synthesis is a compelling and valuable research avenue.

In this paper, we propose \textbf{CityGenAgent}, a natural language-driven framework for hierarchical procedural generation of high-quality 3D cities. At the core of our approach are two domain-specific language (DSL) programs, the \textbf{Block Program} and the \textbf{Building Program}, which provide a two-level decomposition and parameterization of cities. These programs offer a compact yet expressive representation: a city block layout and a building structure encoded by a simple set of parameters, which not only facilitates efficient generation but also enables controllable editing and manipulation.

Built upon these programs, we introduce two specialized agents, \textbf{BlockGen} and \textbf{BuildingGen}, trained via Supervised Fine-Tuning (SFT) and Reinforcement Learning (RL). In the SFT warm-up stage, the model learns basic instruction-following capabilities, ensuring correct program formatting. To mitigate data scarcity and prevent memorization of SFT patterns, RL is applied in the post-training phase to improve generalization. We propose \textbf{Spatial Alignment Reward} and \textbf{Visual Consistency Reward} to enhance spatial reasoning and align outputs with human preferences.
BlockGen focuses on generating coherent block layouts, placing buildings and urban elements in physically plausible ways that avoid collisions and maintain appropriate density. Its Spatial Alignment Reward combines rule-based metrics with human preference considerations to produce spatially correct and logically coherent Block Program. BuildingGen, in contrast, generates Building Program whose rendered appearances faithfully reflect textual specifications. The Visual Consistency Reward evaluates alignment with input conditions, including facade details, style, and materials, guiding BuildingGen to produce semantically accurate and visually cohesive city buildings.

By employing programs as editable proxies and designing the reward to enhance generalization capabilities, our system further enables fine-grained control over city elements. Users can directly modify blocks or buildings through natural language commands, including changes to style, structure, and spatial distribution, without relying on external tools or plugins. 
In summary, our main contributions are:
\begin{itemize}[leftmargin=*, topsep=3pt, partopsep=0pt]
\item We propose programs specifically for 3D city generation, Block Program and Building Program. This approach decomposes city into blocks, buildings, and building components, enabling flexible control and executed.
\item We propose CityGenAgent, consisting of BlockGen and BuildingGen. By introducing Spatial Alignment Reward and Visual Consistency Reward, we enhance the model's spatial reasoning and ensure coherent visual fidelity.
\item Experimental results demonstrate that CityGenAgent is capable of accurately following user instructions to generate high-quality 3D cities. Furthermore, the system supports users to interactive manipulate the block and building by natural language.
\end{itemize}
\begin{figure*}[t]
    \centering
\includegraphics[width=\textwidth]{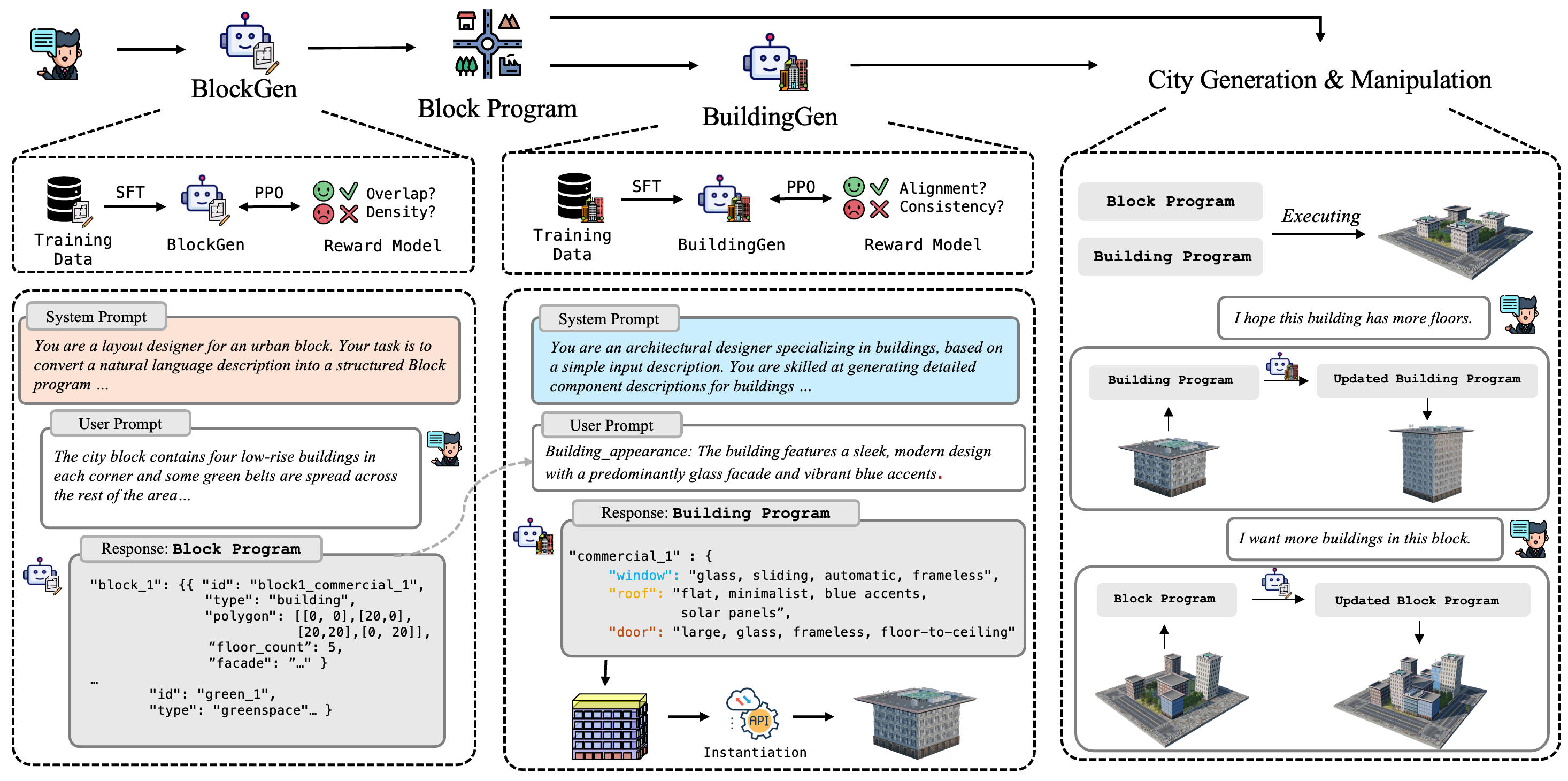}
 \caption{Overview. \textbf{BlockGen} (left) converts user prompt into structured Block Program that defines spatial layouts of urban elements.
 \textbf{BuildingGen} (middle) refines each block by producing Building Program that captures architectural attributes.
\textbf{Block Program} and \textbf{Building Program} are then executed into 3D city instances (right), which can be interactively manipulated via natural language refinement.
 }
    \label{fig:overview}
    \vspace{-3.5ex}
\end{figure*}
\section{Related Work}
\subsection{Rendering and Diffusion-based Scene generation.}
Scene generation is typically addressed through rendering-based and diffusion-based methods.
Neural rendering-based approaches~\citep{chen2023scenedreamer,lin2023infinicity, xie2024citydreamer, 10.5555/3600270.3601875_SGAM} use implicit representations of 3D scenes and apply volumetric rendering to neural fields. For instance, CityDreamer~\citep{xie2024citydreamer} segments the urban environment into buildings and backgrounds, employing distinct neural field types. While these methods achieve impressive visual quality, their lack of 3D geometric fidelity and user control restricts their applicability in downstream tasks. Some research has increasingly explored the use of diffusion-based methods to generate layouts or scenes~\citep{inoue2023layoutdm, wu2024blockfusion, yu2024wonderjourney, ren2024xcube, bian2024dynamiccity, liu2023exim}. DynamicCity \cite{bian2024dynamiccity} employs the voxel-based representation for large-scale city generation. However, this approach inherently lacks fine-grained geometric details and fails to capture the high-fidelity structural intricacies characteristic of real-world. CityGen~\citep{deng2025citygen} introduces an end-to-end framework capable of generating diverse city layouts using Stable Diffusion but demonstrates limited extensibility in accommodating conditional inputs for subsequent editing and refinement operations.

\subsection{Procedure-based Scene Generation.} Traditional approaches~\citep{10.1145/383259.383292_cityenigne, inproceedings, benevs2014procedural} have established rule-based approaches to generate road networks and buildings, which demand extensive manual modeling and substantial labor costs. Recently, methods like Infinigen~\citep{lee2024infinigen} and Infinigen Indoors~\citep{raistrick2024infinigen} introduce comprehensive procedural systems for generating natural landscapes and indoor scenes through stochastic or constrained mathematical algorithms, yielding highly diverse and photorealistic outcomes.
With the development of LLMs~\citep{ouyang2022training}, researchers have made attempts to sythesis scenes conditioned on user input, including general scenes~\citep{zhang2025scene, gao2024graphdreamer, zhou2024scenex, sun20253d, liu2025worldcraft} and indoor scenes~\citep{feng2023layoutgpt, sun2025layoutvlm, fu2024anyhome}.
In city generation, Yo’City~\citep{lu2025yo} and MajutsuCity~\citep{huang2025majutsucity} achieve high fidelity and stylistic adaptability but rely on large-scale 3D assets or multi-stage pipelines and offer limited spatial reasoning capabilities. CityCraft~\citep{deng2024citycraft} and UrbanWorld~\citep{shang2024urbanworld} lack explicit structural definitions and depend on pre-existing priors, restricting diversity and scalability. CityX~\citep{zhang2024cityx} uses a multi-agent framework but depends heavily on contextual plugin coordination, hindering reliable pipeline execution. A summary is provided in Table~\ref{summary_city}.
\section{Method}
\label{sec:Method}
\subsection{Overview}
Given an input description of a city block $I$, our goal is to generate a visually coherent and semantically consistent 3D city block $H$. 
Figure~\ref{fig:overview} provides an overview of our framework. BlockGen and BuildingGen are finetuned based on LLMs in two stages: SFT on instruction–program pairs, followed by Proximal Policy Optimization (PPO)~\citep{schulman2017proximal} to enhance spatial reasoning and visual consistency. Both programs serve as editable intermediates that can be executed to assemble the final 3D city, and more importantly, enable manipulation of the generated city.
\subsection{BlockGen}
\label{blockgen}
\paragraph{Goal and interface.}
{Given a description $I$ of a city block, BlockGen outputs Block Program $P_{\text{block}}$ that parameterizes the block layout, including the placement and attributes of buildings and other elements like greenspaces.
BlockGen is trained to map user instructions into Block Program and we refine it through SFT and PPO to enhance spatial reasoning.
\paragraph{Block Program.}
A Block Program $P_{\text{block}}$ encodes the block layout as an ordered list of elements \(P_{\text{block}}=\langle b_1,\ldots,b_n\rangle\), where each element \(b\) contains the following fields. The first three fields are required, while the last two are optional and apply only to buildings. An example Block Program is provided in Appendix~\ref{example for block program}.

\begin{itemize}[leftmargin=*, itemsep=1pt, parsep=0pt, topsep=0pt]
\item \texttt{id} (string, required): A unique identifier for the element.
\item \texttt{type} (string, required): The usage category of the element, such as \texttt{"residential"}.
\item \texttt{polygon} (list of $[x,y]$, required): A simple (non-self-intersecting) footprint represented as a counter-clockwise ordered list of 2D vertices in block coordinates (meters).
\item \texttt{floor\_count} (integer $\geq 1$, optional): The number of floors for a building.
\item \texttt{facade} (string, optional): A natural-language description of the building’s facade appearance.
\end{itemize}

\subsubsection{BlockGen Supervised Fine-Tuning (Block-SFT)}
During the SFT cold-start stage, the model learns to follow instructions and produce outputs in the correct format, ensuring complete fields and geometrically closed shapes. To support training, we construct a paired dataset in which each sample consists of an input prompt and its corresponding target block layout. The raw data pairs are further post-processed to remove low-quality samples, shown in Appendix \ref{Dataset Processing}. 
In this process, BlockGen is trained to generate the valid Block Program, capturing basic spatial relationships among block elements and aligning them with user instructions.

\subsubsection{Spatial Alignment Reward Preference Optimization (BLOCK-PPO)}

Simple SFT on our limited synthetic dataset reliably teaches BlockGen to produce well-formed block programs, but does not yield robust spatial reasoning or generalization to complex, unseen scenarios. We therefore design specific rewards and adopt RL to enhance the spatial reasoning of our BlockGen.
Concretely, we define \textbf{Spatial Alignment Reward} that scores each generated Block Program from two complementary perspectives: \textbf{Semantic Consistency}, which measures its consistency to the input descriptions $I$, e.g., correct types and relative placements, and \textbf{Spatial Structural Consistency}, which encourages physically plausible layouts, e.g., non-overlapping footprint. 
This reward evaluation allows the model to move beyond SFT’s format alignment and learn policies that generalize to more complex and even out-of-distribution layouts.

\textbf{Semantic Consistency Evaluation.} The evaluation is to quantify how well a predicted Block Program semantically aligns with the user instruction $I$. Measuring semantic alignment is challenging since text provides high-level instructions while the Block Program specifies low-level geometry. To bridge this gap, we render the program as a 2D image and use GPT-4o~\citep{achiam2023gpt} to assess semantic alignment and global plausibility with a standardized prompt (see Appendix~\ref{Semantic Consistency Evaluation}) to obtain two scalar scores: 
\begin{itemize}[leftmargin=*, itemsep=1pt, parsep=0pt, topsep=0pt]
    \item Semantic Alignment $S_{\text{align}}\in[0,10]$, assessing faithfulness of the layout to the user input.
    \item Global Plausibility $S_{\text{plau}}\in[0,10]$, assessing whether the arrangement is physically plausible.
\end{itemize}

\textbf{Spatial Structural Consistency Evaluation.} 
Beyond semantic alignment and global plausibility, good urban layouts should (i) avoid overlap of elements and (ii) maintain a reasonable built-area coverage. 
For the given block program, we therefore introduce two simple but broadly applicable priors: Geometric Overlap and Footprint Density, measuring the interpenetration and building-area coverage of the block program to form our spatial objective.
\begin{itemize}[leftmargin=*, itemsep=2pt, parsep=0pt, topsep=0pt]
\item Geometric Overlap (S\textsubscript{overlap}) :

Given a Block Program $P_{\text{block}}=\langle b_1,\dots,b_n\rangle$, each building $b_i$ stores fields such as \texttt{id}, \texttt{type}, and \texttt{polygon}. We use only the \texttt{polygon} for geometric overlap. Let $L$ be the area of the block region, $A(\cdot)$ the area operator and $R_i$ denote the area of the $i$-th \texttt{polygon}. For each $b_i$, let
{
\setlength{\abovedisplayskip}{5pt} 
\setlength{\belowdisplayskip}{5pt} 
\begin{align}
\mathrm{poly}(b_i)=\langle (x_{i,1},y_{i,1}),\dots,(x_{i,m_i},y_{i,m_i})\rangle,
\end{align}
}be its simple (non-self-intersecting) polygon. The axis-aligned bounding box of $b_i$, denoted as $R_i = [x_i^{\min},x_i^{\max}] \times [y_i^{\min},y_i^{\max}]$, is determined by the extremal coordinates of its vertices: $x_i^{\min}$ and $x_i^{\max}$ are the minimum and maximum x-coordinates, while $y_i^{\min}$ and $y_i^{\max}$ are the minimum and maximum y-coordinates among all its vertices. We then define the geometric overlap percentage $O$ 
as the ratio between the total pairwise intersection area of all bounding boxes 
and the area of the layout region $L$. Specifically, we sum the intersection 
areas $\mathrm{A}(R_i \cap R_j)$ over all unordered pairs $(i, j)$ with $i < j$, 
and normalize this total by the area $\mathrm{A}(L)$ of the entire layout.
The overlap percentage \(O\) is normalized to a 0--10 scale, yielding the spatial overlap score \(S_{\text{overlap}} = 10 \times (1 - O)\). A score of 10 indicates no overlap, with the score decreasing as overlap increases.

\item Footprint Density (S\textsubscript{density}) :

To complement $S_{\text{overlap}}$, we assess built-area coverage against a target density band $[D_{\min}, D_{\max}]$ using the same bounding box ${R_i}$. We define a density score $S_{\text{density}}$ to encourage building coverage within this band. Layouts within the band receive higher scores, while scores decrease proportionally for under- or over-coverage. In our experiments, we set $D_{\min} = 0.5$ and $D_{\max} = 0.8$ to balance efficiency and practicality.

\end{itemize}

\textbf{Spatial Alignment Reward.}
We define the final Spatial Alignment Reward as the mean of four reward scores:
semantic alignment ($S_{\text{align}}$), global plausibility ($S_{\text{plau}}$), geometric
overlap ($S_{\text{overlap}}$), and footprint density ($S_{\text{density}}$):
\vspace{-4pt}
\begin{align}
    S_{\text{spatial}}
    = \frac{1}{|\mathbb{S}|} \sum_{S_i \in \mathbb{S}} S_i,
\end{align}
\vspace{-2pt}Here, $\mathbb{S} = \{ S_{\text{align}},\; S_{\text{plau}},\; S_{\text{overlap}},\; S_{\text{density}} \}$. This simple averaging keeps contributions balanced without extra hyperparameters; a weighted variant can be used if different priorities are desired.

\textbf{Preference Optimization.}
We use PPO to enhance the spatial reasoning ability of BlockGen. Following the common practice~\citep{ouyang2022training}, we construct the preference pairs for training the reward model to predict a scalar score that reflects the target preference. Then we use the output of the
reward model as the reward signal to supervise the policy model.

\subsection{BuildingGen}
\label{buildinggen}
\paragraph{Goal and interface.}
Given the building facade description in \texttt{facade} key of Block Program, BuildingGen is trained to map the description to Building Program $P_{\text{building}}$, which decomposes the building into distinct components and provides detailed feature for each component. 
We fine-tune BuildingGen on LLMs via SFT and PPO for semantic alignment and visual consistency.

\paragraph{Building Program.}
A Building Program $P_{\text{building}}$ encodes the appearance of a building into the description of its components \(P_{\text{building}}=\langle c_1,\ldots,c_n\rangle\), where each component \(c\) has two required fields, defined as follows, with an example shown in Appendix \ref{example for building program}.
\begin{itemize}[leftmargin=*, itemsep=1pt, parsep=0pt, topsep=0pt]
\item \texttt{type} (string): A category to describe the usage of the component, such as \texttt{"door"}.
\item \texttt{description}: A natural language, consisting of several phrases, describing the component's color, style, material, and other decorative details, such as \texttt{"large, blue, frameless"}.
\end{itemize}

\begin{figure*}[t]
    \centering
    \includegraphics[width=0.85\textwidth]{
    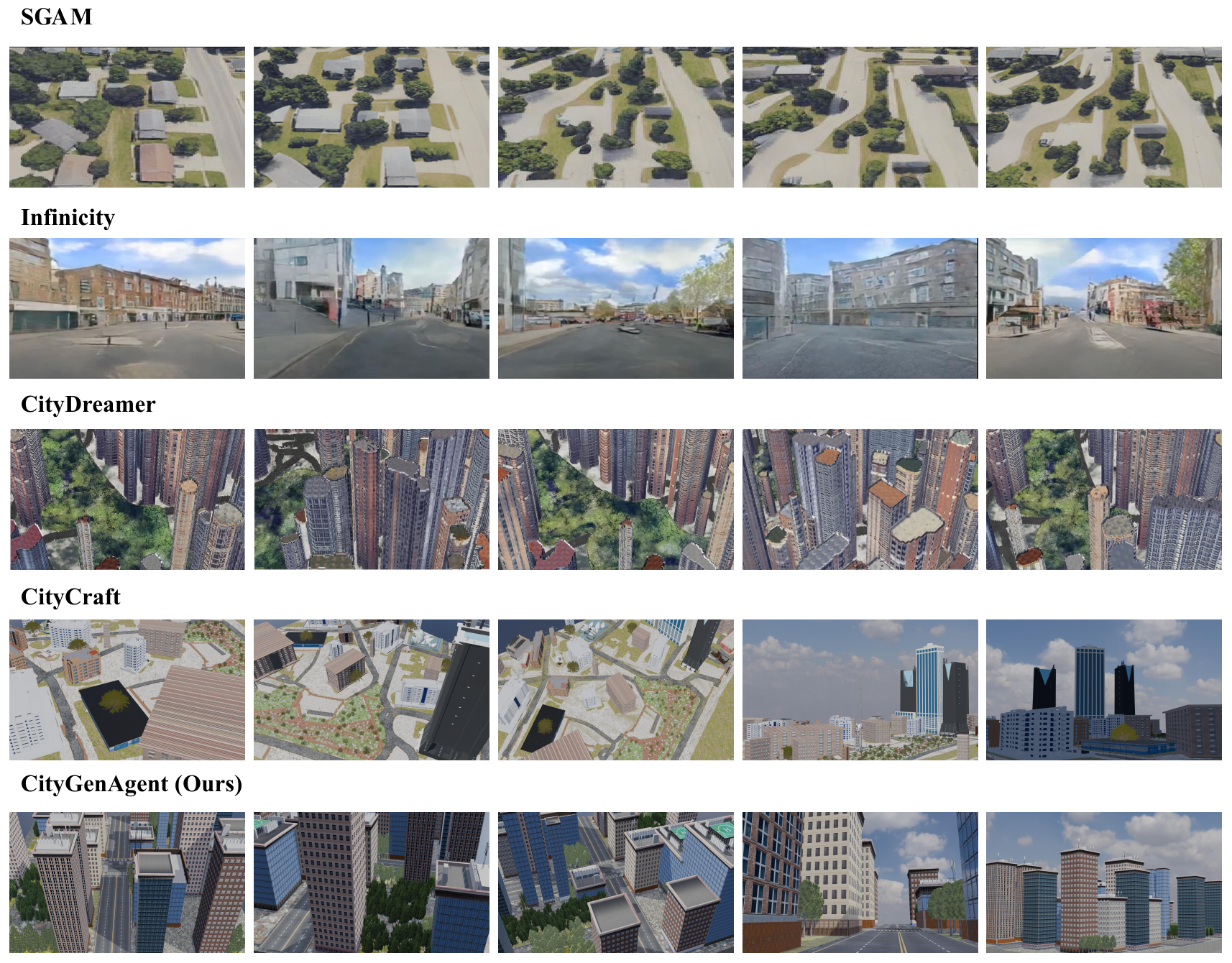}
\caption{Comparison Results of City Generation.}
  \label{comarison with existing work}
    \vspace{-3.5ex}
\end{figure*}

\subsubsection{BuildingGen Supervised Finetuning (Building-SFT)} 
\label{3.3.1}
Similar to BlockGen, we use SFT to warm the LLMs.
To achieve this, we construct a synthetic dataset of paired samples for SFT. Each pair consists of a building appearance description and its corresponding component-based Block Program. This step enables BuildingGen to acquire basic format mapping and semantic understanding capabilities, which are essential for subsequent capability enhancement and generalization.

\subsubsection{Visual Consistency Reward Preference Optimization (Building-PPO)}
A modality gap remains after Building-SFT: a text-only model lacks visual grounding, so program text may not match the rendered appearance. We address this with \textbf{Visual Consistency Reward}: execute the generated Building Program to get the renderings, use a VLM to score the visual consistency, train the reward model on these scores, and optimize the policy model using PPO with the reward signal. This closes the gap and steers the model toward programs that render faithfully to the prompt.

\textbf{Visual Consistency Reward.} We develop the visual criteria to assess Visual Consistency Reward score of the rendered buildings based on these key aspects.
\begin{itemize}[leftmargin=*, itemsep=1pt, parsep=0pt, topsep=0pt]
\item Text Alignment: Examines the alignment between the visual result and the input prompt.
\item Color Coherence: Assesses whether the color scheme across the building is harmonious.
\item Style Consistency: Evaluates the consistency of architectural styles among components.
\item Material Coherence: Focuses on the compatibility of materials used throughout the facade.
\end{itemize}

\textbf{Preference Optimization.}
Following the workflow in BlockGen, we use 
the defined visual criteria to construct the dataset to train the reward model and policy model.
We carefully designed the evaluation prompt based on these four criteria. Further training details are provided in Appendix \ref{Visual Consistency Evaluation} and \ref{Training Details}.

\begin{table*}[t]
\caption{Quantitative Comparison on Text Alignment and Visual Consistency.}
\label{comparison}
   \vspace{-0.5ex}
\begin{center}
\begin{tabular}{lcccccccc}
\hline

\multirow{2}{*}{\rule{0pt}{20pt}\textbf{Method}} & \multicolumn{3}{c}{
\multirow{2}{*}{\begin{tabular}{c}
     \textbf{Text} \\
     \textbf{Alignment} 
\end{tabular}$\uparrow$}} &  \multicolumn{2}{c}{\multirow{2}{*}{\begin{tabular}{c}
     \textbf{Visual} \\
     \textbf{Consisitency} 
\end{tabular}$\uparrow$}}&
\multicolumn{2}{c}{\multirow{2}{*}{\begin{tabular}{c}
     \textbf{Geometric} \\
     \textbf{Quality} 
\end{tabular}}} \\ 
\\
\cmidrule(lr){2-4} \cmidrule(lr){5-6} \cmidrule(lr){7-8}
& CLIP & GPT & User & GPT & User & ROS$\uparrow$ & OTR$\downarrow$\\
\hline
SGAM~\citep{10.5555/3600270.3601875_SGAM} & 0.106  & 3.0  & 1.7 & 5.1  & 4.2 & - & - \\
Infinicity~\citep{lin2023infinicity} & 0.249  & 5.5  & 3.6 & 4.0 & 2.9 & - & -  \\
CityDreamer~\citep{xie2024citydreamer}         & 0.210  & 5.2  & 5.2 & 6.0 & 4.1 & -  & -  \\ \hline
CityCraft~\citep{deng2024citycraft}          & 0.266  & 6.0  & 4.5 & 6.1 & 5.1 & 0.309 & 192.301 \\
Hunyuan3D~\citep{zhao2025hunyuan3d}       & 0.272  & 5.2 & 3.9 & 6.5 & 5.5 & 0.182  & 6999.983 \\
CityGenAgent(Ours)   & \textbf{0.286}  & \textbf{6.6} & \textbf{6.1} & \textbf{6.7} & \textbf{5.8} & \textbf{0.357} & \textbf{177.970} \\
\hline
\end{tabular}
\end{center}
    \vspace{-2.5ex}
\end{table*}
\subsection{Program Execution and Asset Assembly}
\label{sec3.5}
With the Block Program and Building Program, our executor generates 3D city scenes in two stages. 

\textbf{Asset Preparation.} We parse the Block Program to extract building footprints and floor counts, construct base meshes, and use Building Program component descriptions to retrieve assets from our architectural database via semantic matching. To overcome the limitations of a fixed asset set, we also explore Text-to-3D generation (Hunyuan3D~\citep{zhao2025hunyuan3d}) to dynamically expand the component database. Geometric attributes from the Block Program are used to compute placement parameters. For each polygon edge, we derive its length, direction, and outward normal to determine the transformation matrices of attached components.

\textbf{Asset Assembly.} Prepared assets are instantiated and placed through rotation, translation, and scaling to form complete buildings. Additional scene elements, such as roads, trees, and streetlights, are generated according to spatial specifications in the Block Program.
Acting as an intermediary layer, the executor translates program instructions into commands compatible with graphics engines, enabling automated and scalable generation of complex 3D city scenes from text.

\subsection{Interactive Manipulation via Language }
\label{sec3.6}
Leveraging the effective representations of Block Program and Building Program, along with the model's generalization capability acquired through RL, our framework enables users to manipulate individual blocks or buildings via natural language commands. Given the current Block Program or Building Program, the user can provide instructions to the corresponding module, BlockGen or BuildingGen, which then updates the program accordingly to follow the desired changes. For example, BlockGen can modify block density or adjust building heights, while BuildingGen can alter architectural details such as the style of windows and doors. 

\section{Experiments}
\label{sec:Experiments}
\subsection{Experimental Details}
\label{sec:Experimental Details}
\textbf{Dataset.} We construct supervised and preference datasets for both BlockGen and BuildingGen, including 5k SFT pairs and 5k preference pairs for BlockGen, and 5k SFT pairs with 5k preference samples for BuildingGen, as detailed in Appendix~\ref{Dataset Processing}.
For evaluation, we gather 100 city block descriptions and 50 manipulation prompts, subsequently input into CityGenAgent to generate 3D scenes for quantitative and qualitative comparisons. 

\textbf{Model Training.} In our framework, BlockGen and BuildingGen are both finetuned from Qwen3-8B~\citep{yang2025qwen3}. More details are provided in Appendix \ref{Training Details}. 

\textbf{Metrics.}
We evaluate rendered 3D city scenes using \emph{Text Alignment} and \emph{Visual Consistency} assessed by GPT and a user study, and also report CLIP scores~\citep{radford2021learning}. Geometric mesh quality is measured by two metrics: ROS for edge orthogonality and OTR for tessellation efficiency (see Appendix~\ref{gpt_eval}).
For BlockGen, we adopt three layout-level metrics: \emph{Collision Rate} (Collision), defined as the ratio of total pairwise overlap area to block area, and \emph{Positional Coherency} (Pos.) and \emph{Physics-based Semantic Alignment} (PSA), following LayoutVLM~\citep{sun2025layoutvlm}.
For BuildingGen, GPT-4o is used to evaluate building-level \emph{Text Alignment} and \emph{Visual Consistency}. To further assess program validity, we introduce \emph{Format Accuracy}, which measures (i) JSON parsability, (ii) geometric validity of polygon definitions, and (iii) completeness of required fields.

\begin{figure*}[h]
   \centering
   \includegraphics[width=\textwidth]{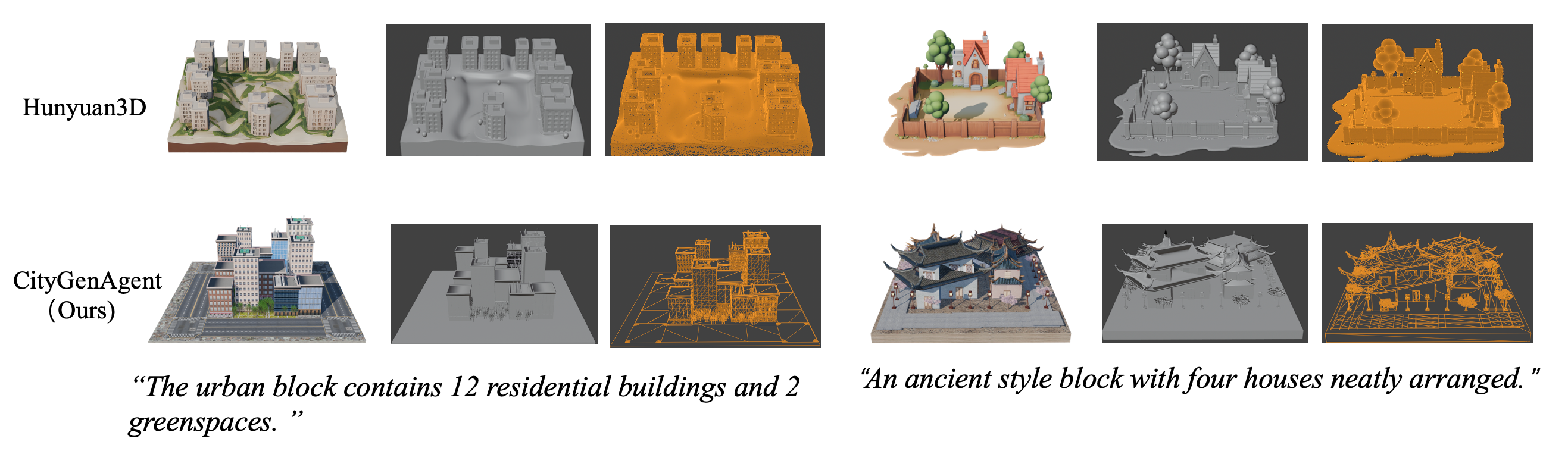}
 \caption{
Qualitative Comparison with Hunyuan3D. We present the prompts, rendered images, mesh visualization, and wireframe visualization for each scene.}
 \label{hy_vs}
    \vspace{-2ex}
\end{figure*}

\subsection{Comparison with existing methods}
\textbf{Quantitative Comparison.}
As shown in Table~\ref{comparison}, our method consistently outperforms existing city generation approaches in both \textit{Text Alignment} and \textit{Visual Consistency}. CityGenAgent achieves the highest scores across the text-alignment metrics and visual-consistency evaluations, indicating stronger semantic faithfulness to the input description and more coherent rendered results. 

In terms of geometric quality, CityGenAgent produces meshes with better rectilinearity, achieving the highest ROS score and lowest OTR among all methods. Compared with Hunyuan3D, CityGenAgent lowers OTR by nearly 40×, and also improves over CityCraft, demonstrating that our procedural generation strategy leads to more regular structures and a more efficient distribution of mesh elements. As shown in Table~\ref{compliance metric}, CityGenAgent further achieves the best overall performance on program-level compliance metrics, outperforming all compared LLMs (GPT-4o~\citep{achiam2023gpt}, Qwen2.5-7B~\citep{team2024qwen2} and Qwen3-8B~\citep{yang2025qwen3}). It maintains high format accuracy while keeping collision rates low, which is essential for generating structurally valid urban layouts. The RL stage improves spatial reasoning without sacrificing program validity, leading to better collision avoidance and semantic consistency. These results validate the effectiveness of our reward design in guiding the model toward structurally sound and semantically aligned city layouts.

\begin{table*}[h]
\centering
\caption{Ablation Study. We evaluate different RL methods for BlockGen and BuildingGen.}
\label{ablation_table}
\begin{adjustbox}{max width=0.9\textwidth}
\begin{tabular}{@{}c c@{}}
\subcaptionbox{BlockGen\label{tab:blockgen}}{
\begin{tabular}{c|ccc}
\textbf{Method} & \textbf{Collision} $\downarrow$ & \textbf{Pos.} $\uparrow$ & \textbf{PSA} $\uparrow$ \\
\hline
Base Model         & 23.97\% & 76.70 & 75.60 \\
Base Model + SFT   & 5.59\%  & 80.17 & 84.02 \\
Base Model + DPO   & 5.19\%  & 81.13 & 85.03 \\
Base Model + PPO   & \textbf{4.89\%} & \textbf{85.33} & \textbf{87.90} \\
\end{tabular}
}
&
\subcaptionbox{BuildingGen\label{tab:buildinggen}}{
\begin{tabular}{c|cc}
\textbf{Method} & \textbf{Text Alignment} $\uparrow$ & \textbf{Consistency} $\uparrow$ \\
\hline
Base Model         & 5.5 & 5.7 \\
Base Model + SFT   & 6.8  & 8.7  \\
Base Model + DPO   & 7.0  & 8.1  \\
Base Model + PPO   & \textbf{7.5} & \textbf{8.9} \\
\end{tabular}
}
\end{tabular}
\end{adjustbox}
  \vspace{-2.2ex}
\end{table*}


\textbf{Efficiency Evaluation.}
We compare CityGenAgent with Hunyuan3D~\citep{zhao2025hunyuan3d}, CityCraft~\citep{deng2024citycraft}, and manual modeling by PCG experts on an NVIDIA H100 NVL GPU. CityGenAgent generates a single block (0.75 min) much faster than Hunyuan3D (3 min), CityCraft (1 min), and manual modeling (60 min). As for token efficiency, CityGenAgent achieves a 17.7\% performance gain (91.59 vs. 77.83) with only 4.1\% additional tokens (1134 vs. 1089), yielding a 13.0\% improvement in token efficiency (8.08 vs. 7.15), as detailed in Appendix~\ref{app:token-efficiency}.

\begin{table}[h]
\caption{Quantitative Comparison of Different Language Models in City Generation.}
\label{compliance metric}
\begin{center}
\resizebox{\linewidth}{!}{
\begin{tabular}{c|cccc}
\hline
\textbf{Method} & 
\makecell{\textbf{Format} \\ \textbf{Accuracy}} & \textbf{Collision} & \textbf{Pos.} & \textbf{PSA}  \\
\hline
GPT-4o & 70\%   & 6.67\% & 78.45 & 85.10\\
Qwen2.5-7B & 70\%   & 37.99\% & 67.60 & 61.25\\
Qwen3-8B  & 83\%  & 23.97\% & 76.70 & 75.60 \\
\noalign{\vskip 2pt} 
\hline
\noalign{\vskip 2pt} 
CityGenAgent w/o RL  & 98\% & 5.59\% & 80.17 & 84.02  \\
CityGenAgent & \textbf{98\%}  & \textbf{4.89\%} & \textbf{85.33} & \textbf{87.90} \\
\hline
\end{tabular}
}
\end{center}
    \vspace{-3.5ex}
\end{table}

\begin{figure*}[t]
    \centering
    \includegraphics[width=1\linewidth]{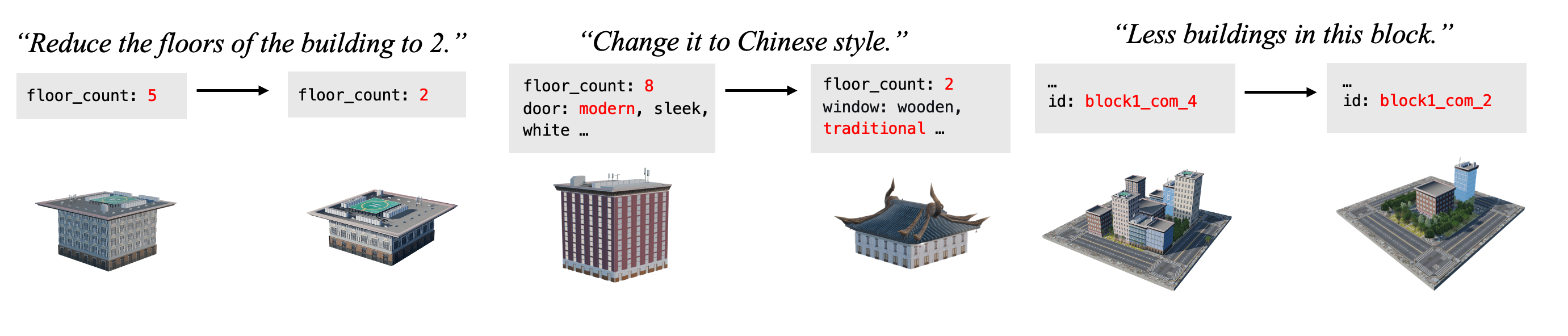}
    \caption{Scene Manipulation Results.} 
    \label{mani_appendix}
        \vspace{-3.5ex}
\end{figure*}

\textbf{Qualitative Comparison.}
As illustrated in Figure \ref{comarison with existing work}, the qualitative results highlight the superiority of CityGenAgent. In comparison, InfiniCity, SGAM, and CityDreamer all suffer from low clarity and a lack of urban details. Although CityCraft demonstrates coherent 3D structures but the scale and harmony between buildings lack consistency and do not conform to real-world rules. 
As illustrated in Figure~\ref{hy_vs}, Hunyuan3D’s renderings are constrained to a cartoon-like style and exhibit limited scalability. Mesh and wireframe visualizations further reveal that it often produces irregular and overly dense tessellations, resulting in redundant geometry. In contrast, our method preserves generative flexibility while producing clean, planar surfaces with significantly improved mesh regularity and structural clarity, leading to superior visual and geometric quality. Additional visual results of CityGenAgent are presented in Appendix~\ref{more_result}.

\subsection{Manipulation}
\label{mani_section}
The city block produced by our framework can be further generalized to be manipulated using natural language, as shown in Figure \ref{mani_appendix}.
Unlike rendering-based approaches that operate in an entangled pixel or latent space, our framework leverages explicit procedural representations, Block Program and Building Program, to decouple scene layout from architectural composition. This structured formulation enables precise, parametric control over the scene while preserving geometric plausibility and aesthetic coherence. By bridging natural language instructions with parametric programs, our system supports multi-granularity interaction: users can employ fuzzy semantic commands (e.g., changing style) or precise numerical constraints (e.g., adjusting floor counts). Notably, our two-stage training paradigm (SFT followed by RL) on synthetic data empowers the model with emergent spatial reasoning capabilities. As evidenced by the ``Change it to Chinese style'' manipulation in Figure \ref{mani_appendix}, the model demonstrates capabilities beyond superficial texture mapping. It successfully performs cross-attribute inference: although the prompt only specifies a stylistic change, the model recognizes the implicit geometric constraints associated with traditional Chinese architecture. Consequently, it autonomously reduces the floor count to align the structural topology with the requested historical context. This demonstrates that CityGenAgent has internalized the joint distribution of architectural style and spatial topology, rather than merely memorizing training patterns.


\subsection{Ablation Study}

To validate the effectiveness of the reward design, we compare the performance of Direct preference optimization (DPO)~\citep{rafailov2023direct} and PPO using our preference sample pairs in the RL stage. The results are shown in Table~\ref{ablation_table} and Figure~\ref{ablation_img}.

\textbf{BlockGen.}
We find that although DPO enhances spatial structure through preference learning, it exhibits limitations in capturing fine-grained geometric constraints when compared with PPO. The reward model trained in the PPO framework enables more effective handling of multi‑dimensional spatial rewards, including collision avoidance and alignment consistency, thereby guiding the model to achieve better overall performance.


\textbf{BuildingGen.
}
BuildingGen exhibits a similar trend. Relative to the base model, SFTand RL consistently improve performance across dimensions. Moreover, under the same training samples, PPO surpasses DPO, indicating greater training stability and a stronger ability to capture human preferences. We further reveal that our designed reward formulation is the necessary condition for defining the correct optimization manifold. By leveraging the RL algorithm to enhance the model's capabilities across target dimensions, our approach yields plausible results that are significantly more aligned with human preferences.
\section{Conclusion}
In this paper, we presented CityGenAgent, a natural language-driven framework for hierarchical procedural 3D city generation. By introducing Block Program and Building Program, we achieved disentangled control over the spatial layout and architectural composition of city elements. BlockGen and BuildingGen are optimized through SFT and RL with tailored reward mechanisms to ensure robust spatial reasoning and visual consistency. Extensive experiments demonstrate that our framework yields results with superior spatial rationality and visual fidelity. CityGenAgent also supports fine-grained manipulation through natural language instructions. This work can establish a foundation for city modeling and interactive content creation. Please refer to Appendix \ref{limitation} for a discussion on limitations.
\section*{Impact Statement}
This work presents a technical framework for structured 3D city generation and does not involve human subjects, personal data, or privacy-sensitive information.  We encourage responsible use and further evaluation of ethical considerations in downstream applications.

\bibliography{city_reference}
\bibliographystyle{icml2026}


\clearpage
\appendix
\section*{Appendix}
\section{Implementation Details}
\subsection{Prompts}
In this section, we provide prompts used for training and testing the model.

\textbf{BlockGen-SFT Data Generation.} In order to enable BlockGen to output the corresponding program in the SFT stage, we use the following prompts to generate Block Program.
\label{BlockGen-SFT Data Generation}
\begin{tcolorbox}[enhanced, breakable, colback=gray!10, colframe=gray!80, boxrule=0.5pt, arc=2pt, left=12pt, right=12pt, top=12pt, bottom=12pt, fontupper=\fontfamily{pcr}\selectfont]

You are a layout parser for urban block descriptions.

Your task is to convert a natural language description of an urban block into a structured JSON layout. 
Each building or green space must be assigned:

    - a unique "id",
    
    - a "type" (such as "resident building" ,"school",      "library", "commerical", building", "office", "greenspace"),
    
    - a "polygon" represented as a list of [x, y] points (for example, x, y are integers within [0, 100]),
    
    - and for buildings,  there is a key "floor count" should be a digital number
    and a key "facade" to  describe the general appearance of the building. \\
         
        Make sure the polygons do not overlap and collectively cover the area proportions mentioned in the description. Keep the shapes simple (e.g., 
        rectangles or L-shapes), and follow the stated quantity and proportions as accurately as possible.\\
        
        Your output must follow **this exact JSON format**:
        \begin{verbatim}
[ 
{
    "description": "<repeat the 
    input description here>",
    "layout": 
    {
    "buildings": [
    {
    "id": "res_1",
    "type": "resident building",
    "polygon": [
    [x1, y1], [x2, y2], 
    [x3, y3], [x4, y4]],
    "floor_count": 16,
    "facade": 
    "glass with wooden accents"
        
    },
    ...
    ]
    "greenspaces": [
    {
    "id": "green_1",
    "type": "greenspace",
    "polygon": 
    [[x1, y1], [x2, y2], 
     [x3, y3], [x4, y4]]
    },
    ...
    ]
    }
}
]
Prompt: 
{Text description}
\end{verbatim}
\end{tcolorbox}

\textbf{BuildingGen-SFT Data Generation.} In order to enable BuildingGen to output the corresponding program in the SFT stage, we use the following prompts to instruct GPT-4o to generate Block Program.
\label{BuildingGen-SFT Data Generation}
\begin{tcolorbox}[enhanced, breakable, colback=gray!10, colframe=gray!80, boxrule=0.5pt, arc=2pt, left=12pt, right=12pt, top=12pt, bottom=12pt, fontupper=\fontfamily{pcr}\selectfont]

You are an architectural designer specializing in buildings. Based on a single frontal image of a building facade, produce structured descriptions suitable for 3D modeling or procedural generation.\\

Obey the following rules:
- Input: one frontal image of the building facade.

- Output must strictly be valid JSON, with no additional text or explanations.

- The JSON is an array containing one object with exactly two top-level keys: "facade" and "output".

- "facade": a single short sentence summarizing the overall facade appearance (mention color, style, material, and key characteristics). Keep it concise and continuous, not a list of words.

- "output": an object with exactly these keys — "window", "door", "roof". Each value must be a concise, comma-separated phrase (style/material/structural
/ornamental/color descriptors; no sentences, no subkeys, no nesting; no trailing commas).

\begin{verbatim}
Here is the output 
format you must follow:
[
  {
"facade": "The facade 
    features a modernist 
    design with 
    light gray concrete
    and large glass panels.",
"output": {
  "window": 
      "rectangular modules,
      aluminum frames, 
      clear glass, 
      repetitive grid, 
      slim mullions",
  "door": 
      "single-leaf, 
      glass panel,
      metal frame,
      flush alignment, 
      minimal handle",
  "roof": 
      "flat slab,
      parapet edge,
      concealed drainage, 
      concrete, 
      clean silhouette"
    }
  }
]

\end{verbatim}
\end{tcolorbox}

\textbf{Semantic Consistency Evaluation.} 
For Block-PPO, we calculate the semantic consistency score of samples to construct the positive and negative sample pairs.
To evaluate the semantic consistency of the Block Program sample, we use the following prompt as input to GPT-4o.
\label{Semantic Consistency Evaluation}
\begin{tcolorbox}[enhanced, breakable, colback=gray!10, colframe=gray!80, boxrule=0.5pt, arc=2pt, left=12pt, right=12pt, top=12pt, bottom=12pt, fontupper=\fontfamily{pcr}\selectfont]

You are an urban planning expert and are asked to analyze and score the city block layout image based on the text description. Please strictly follow the following criteria and output the result in a structured JSON format. 

Blue represents buildings, and green represents green space.
\\

Please refer to the following criteria:

1. **Semantic Alignment**: Does the layout conform to the 
text description, especially regarding quantity, distribution, and orientation?

2. **Global Plasusibility**: Does the overall layout exhibit physical feasibility and structural coherence? Are the spatial arrangements consistent with real-world constraints such as accessibility and functional organization?

\begin{verbatim}
Please:
- Read and analyze the image file;
- Assign an integer score 
(0 to 10) to each dimension;
- Save the results in the 
following JSON format.

The output format is as follows:

{
    "image_1": 
    {
        "semantic_alignment": 6,
        "global plasusibility": 8
    }
}

Text description:
{text_description}
Block_layout:
{block_layout_img}
\end{verbatim}
\end{tcolorbox}
\textbf{Visual Consistency Evaluation.} For Building-PPO, we evaluate the visual consistency score of the rendered result to obtain the preference data pairs to train the reward model. We use the following prompt to evaluate the visual results of the executed Building Program.
\label{Visual Consistency Evaluation}
\begin{tcolorbox}[enhanced, breakable, colback=gray!10, colframe=gray!80, boxrule=0.5pt, arc=2pt, left=12pt, right=12pt, top=12pt, bottom=12pt, fontupper=\fontfamily{pcr}\selectfont]
You are given a facade view image of a building. Please evaluate the image from the following perspectives:

- Text Alignment: Examine the degree of correspondence between the completed facade design and the text description provided. Consider whether the overall style, material, color scheme, and structural features match the description.

- Color Coherence: Assess whether the color scheme across the facade is harmonious and visually consistent.

- Style Consistency: Evaluate the consistency of architectural styles among components such as windows, doors, and roofs.

- Material Coherence: Focus on the compatibility and uniformity of materials used throughout the facade.

For each aspect, provide:

- A score from 1 to 10 (1 = very poor, 10 = excellent)

- A brief explanation for your rating

**Output format:**
\begin{verbatim}
Text Alignment: 
[score]/10 — [explanation]
Color Coherence: 
[score]/10 — [explanation]
Style Consistency: 
[score]/10 — [explanation]
Material Coherence: 
[score]/10 — [explanation]

Text description:
{Text_description}
Input image:
{Input_img}
\end{verbatim}

\end{tcolorbox}

\section{Evaluation} 
\label{gpt_eval}
\textbf{GPT-based Evaluation.} To evaluate the quality of the generated 3D cities, we instruct GPT-4o to give the score from the specific aspects. The prompt is shown as follows. 
\begin{tcolorbox}[enhanced, breakable, colback=gray!10, colframe=gray!80, boxrule=0.5pt, arc=2pt, left=12pt, right=12pt, top=12pt, bottom=12pt, fontupper=\fontfamily{pcr}\selectfont]

You are given a top-down view image and a text description of a city block.
Please evaluate the scene based on the following aspects:\\

Consistency: 

    Color Cohesion: Are the colors of windows, doors, and roofs across different buildings in the block harmonious?
    
    Style Consistency: Do the buildings generally belong to a similar architectural style or era?
    
    Authenticity: Does the block conform to the morphology and layout of real-world urban environments?
    
    Quality: Is the image clear and high-fidelity?

Text Alignment: Does the appearance match the given input description?\\

For each aspect, provide a score from 1 to 10:

0-3: Severe Dissonance (clashing, chaotic, lacking cohesion)

4-6: Neutral/Mixed (partially harmonious but with noticeable inconsistencies)

7-8: Harmonious (cohesive and intentional, with minor acceptable variations)

9-10: Highly Harmonious/Uniform (excellent cohesion; tells a clear, unified design story)\\

A brief explanation for your rating.

Output format:

Consistency: [score]/10 — [explanation]

Text Alignment: [score]/10 — [explanation]

\begin{verbatim}
Text description:
{Text_description}
Input image:
{Input_img}
\end{verbatim}

\end{tcolorbox}
\textbf{User Study Details.}
In our experiments, we employ manual evaluation to assess the generated results. We recruited 70 volunteers to participate in the scoring process. An example of the evaluation interface is shown in Figure \ref{user study}. To ensure fairness and objectivity, all evaluation images were anonymized to eliminate potential bias. This procedure helped maintain the integrity and reliability of the evaluation process.

\begin{figure}[h]
    \centering
    \includegraphics[width=0.5\linewidth]{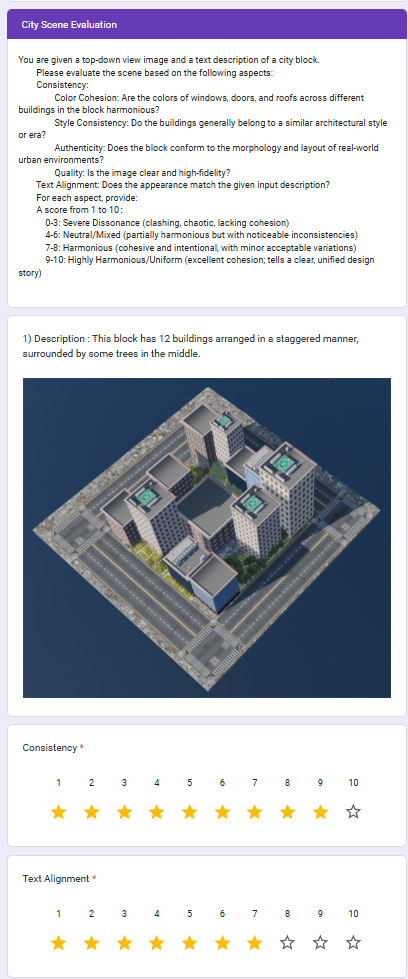}
    \caption{\textbf{Template Questionnaire Used in Participant Studies.} 
}
    \label{user study}
\end{figure}

\textbf{Geometric Quality.} To assess geometric quality of our method, we use the architectural regularity and efficiency indicators. (i) \textbf{Rectilinearity/Orthogonality Score (ROS)} measures the proportion of edge directions aligned with two dominant orthogonal axes, reflecting facade alignment and structural order.  Higher ROS indicates better orthogonality. (ii) \textbf{Over‑tessellation Ratio (OTR)} compares actual triangle density to curvature-based demand, where lower values indicate more efficient tessellation without unnecessary mesh complexity.

\section{Examples of Block program and Building Program}
\textbf{Block Program.} As follows, we provide an example for Block Program.
\label{example for block program}
\begin{tcolorbox}[enhanced, breakable, colback=gray!10, colframe=gray!80, boxrule=0.5pt, arc=2pt, left=12pt, right=12pt, top=12pt, bottom=12pt, fontupper=\fontfamily{pcr}\selectfont]
\begin{verbatim}
{
    "id": "mixed_1",
    "type": "mixed-use building",
    "polygon": 
    [[0, 0], [22, 0],
    [22, 22], [0, 22]],
    "floor_count": 12,
    "facade": "modern glass 
    and steel with 
    terracotta accents"
},
{
    "id": "mixed_2",
    "type": "mixed-use building",
    "polygon": 
    [[25, 0], [47, 0], 
    [47, 22], [25, 22]],
    "floor_count": 10,
    "facade": 
    "concrete with greenery 
    on the upper floors"
},
... 
{
    "id": "park_1",
    "type": "greenspace",
    "polygon": 
    [[36, 50], [55, 50], 
    [55, 67], [36, 67]]
},
{
    "id": "park_2",
    "type": "greenspace",
    "polygon":
    [[36, 71], [55, 71], 
    [55, 89], [36, 89]]
},
... 
\end{verbatim}

\end{tcolorbox}

\textbf{Building Program.} As follows, we provide an example for Building Program.
\label{example for building program}
\begin{tcolorbox}[enhanced, breakable, colback=gray!10, colframe=gray!80, boxrule=0.5pt, arc=2pt, left=12pt, right=12pt, top=12pt, bottom=12pt, fontupper=\fontfamily{pcr}\selectfont]
\begin{verbatim}
{
  "window": 
      "expansive, glass,
      modern, blue-tinted",
  "door": 
      "sleek, modern, 
      glass, automatic",
  "roof": 
      "flat, sleek, 
      modern, weather-resistant"
}
\end{verbatim}

\end{tcolorbox}

\section{Dataset Constructing}
\label{Dataset Processing}
\textbf{Training Dataset For BlockGen.}
For BlockGen, we curate 5k valid samples for the SFT stage. 
The raw pairs are post-processed to remove invalid or low-quality samples. Concretely, we (i) verify each \texttt{polygon} is a closed simple loop vertices and counter-clockwise ordering; (ii) reject pairs with overlapping between polygons (edge/vertex touching is allowed); and (iii) enforce \emph{appropriate density} so blocks are neither empty nor overfilled. 
For training the reward model in RL, we construct 5k preference pairs, filtered to ensure a reward difference of at least 5 on a 0–10 scale. 

\textbf{Training Dataset For BuildingGen.}
For BuildingGen, since there are few building datasets suitable for this task, we construct a paired dataset of 5k examples. Each pair consists of a natural language building description and its corresponding procedural program. To synthesize this dataset, we collected 5k frontal building images from Google Maps and designed specialized prompts for GPT-4o to generate both holistic descriptions of the building facade and component-level descriptions based on our predefined architectural categories such as door, window, and roof. For training the reward model, we gather 5k diverse prompts and generate 5 samples for each prompt. Each sample was assigned an $S_{\text{visual}}$ score through the aforementioned evaluation process. Training pairs were constructed by selecting sample pairs whose reward difference exceeded a predefined threshold, with the higher-scoring sample labeled as ``chosen'' and the lower-scoring as ``rejected''.

\section{Training Details}
\label{Training Details}
For BlockGen and BuidlingGen, we both adopt a two-stage training pipeline to fine-tune the Qwen3-8B model. Low-Rank Adaptation (LoRA) with a rank of 8 is applied to all target modules. The model is trained for 3 epochs with a batch size of 1, a gradient accumulation steps of 8, and a learning rate of $1 \times 10^{-4}$. A cosine learning rate scheduler with 10\% warm-up is employed, and training is conducted in bfloat16 precision. 

BlockGen’s SFT was conducted on 4×NVIDIA A100 GPUs for approximately 5 hours, followed by RL where the reward model was trained for 10 minutes and the policy model optimized via PPO for 8 hours using the same LoRA and hyperparameter settings. BuildingGen’s SFT was trained on 8×A100 GPUs for about 2 hours, and its RL stage involved 10 minutes of reward model training and 2 hours of PPO optimization, also under the same configuration.


\begin{table}[h]
\centering
\caption{Token Efficiency Comparison}
\label{tab:token-efficiency}
\resizebox{\linewidth}{!}{
\begin{tabular}{lccc}
\hline
\textbf{Model} & \textbf{Tokens} & \textbf{Performance} & \textbf{Efficiency} \\
\hline
Qwen3-8B (Single) & 1089 & 77.83 & 7.15 \\
CityGenAgent & 1134 & 91.59 & 8.08 \\
\hline
\end{tabular}
}
\end{table}

\begin{table}[h]
\centering
\caption{Efficiency Evaluation of City Generation Methods for Per Block (100m x 100m).}
\label{efficiency evaluation}
\resizebox{0.85\linewidth}{!}{
\begin{tabular}{c|c|c}
\hline
\textbf{Method} & \textbf{Block} & \textbf{Inference Time} \\
\hline
human      & 1 block      & 60 min    \\
Hunyuan3D  & 1 block      & 3 min    \\
CityCraft  & 1 block      & 1 min     \\
\hline
\multirow{3}{*}{CityGenAgent} 
           & 1 block  & 0.75 min  \\
           & 4 blocks & 1 min    \\
           & 16 blocks& 3 min  \\
\hline 
\end{tabular}}
    \vspace{-2.2ex}
\end{table}

\section{Efficiency Analysis}
\label{app:token-efficiency}

We conduct a detailed token-efficiency analysis to quantify the trade-off 
between performance gain and computational cost in our multi-agent architecture. 
The total token count includes all inter-agent communications, prompt templates, 
and final output generation.
Table~\ref{tab:token-efficiency} and Table~\ref{efficiency evaluation} present the efficiency comparison between CityGenAgent 
and the single-agent base model. Despite the two-agent framework's 
inherent communication overhead, CityGenAgent uses only 4.1\% more tokens 
(1134 vs. 1089) while achieving 17.7\% higher performance (91.59 vs. 77.83). The marginal increase in token consumption is significantly outweighed by the 
performance improvement, demonstrating that our agent division strategy 
effectively allocates computational resources. 

\section{Manipulation}
\label{mani_appendix_section}
We present GPT and human evaluators with pairs of pre- and post-edit rendered images across 50 samples to assess instruction-following alignment and overall visual consistency. The resulting scores indicate that our editing approach effectively adheres to user instructions while maintaining global visual coherence, as shown in Table \ref{mani_table}.


\begin{table}[h]
\caption{Quantitative Evaluation of Manipulation Performance.}
\label{mani_table}
\begin{center}
\resizebox{0.9\linewidth}{!}{
\begin{tabular}{lccccc}
\hline
\multirow{2}{*}{\rule{0pt}{10pt}\textbf{Method}} & \multicolumn{3}{c}{
\textbf{Text Alignment} } &  \multicolumn{2}{c}{\textbf{Consisitency} }\\
\cmidrule(lr){2-4} \cmidrule(lr){5-6}
& CLIP & GPT & User & GPT & User \\
\hline
Ours  & 0.286  & 8.7 & 8.4 & 6.6 & 5.9 \\
\hline
\end{tabular}
}
\end{center}
\end{table}

\section{More Generated Results}
\label{more_result}
We present more results of buildings and  scenes generated by CityGenAgent in Figure~\ref{visual_result}, \ref{scene_result} and Figure \ref{building_result}.




\section{Limitations}
\label{limitation}
Despite CityGenAgent’s ability to efficiently generate high-quality 3D cities from natural language, it still faces some limitations. First, for large-scale or highly complex scenes, inference time can become significant; while parallel processing helps, achieving real-time online deployment remains challenging and may require further optimization strategies such as model compression, low-bit quantization, or incremental scene generation. Second, deploying the model on mobile or edge devices is constrained by limited compute, memory, and power, necessitating lightweight architectures, pruning, or knowledge distillation, as well as careful integration with real-time rendering pipelines. Addressing these challenges is essential for enabling interactive, real-world applications of CityGenAgent.


\begin{figure*}[h]
    \centering
    \includegraphics[width=1\linewidth]{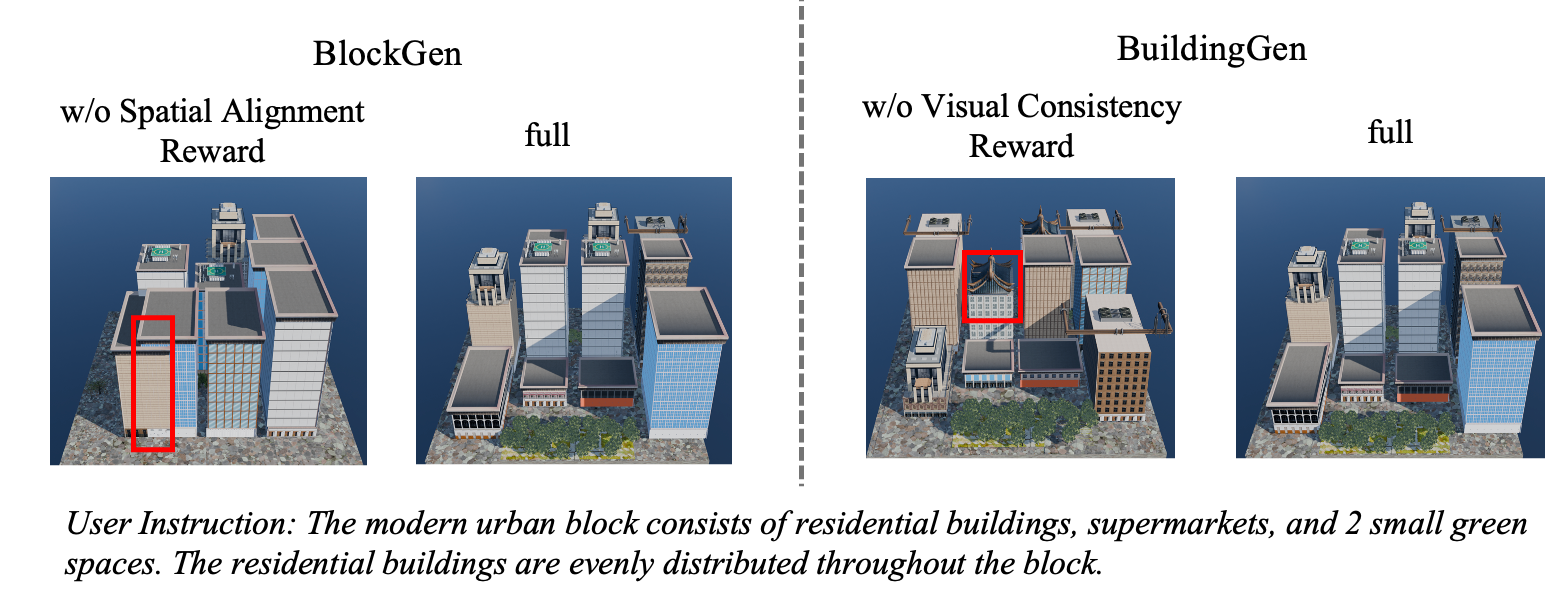}
    \caption{Ablation Study Results. The red boxes highlight the model collision or style mismatch will otherwise occur without reward.
}
    \label{ablation_img}
\end{figure*}

\begin{figure*}[h]
    \centering
    \includegraphics[width=0.7\textwidth]{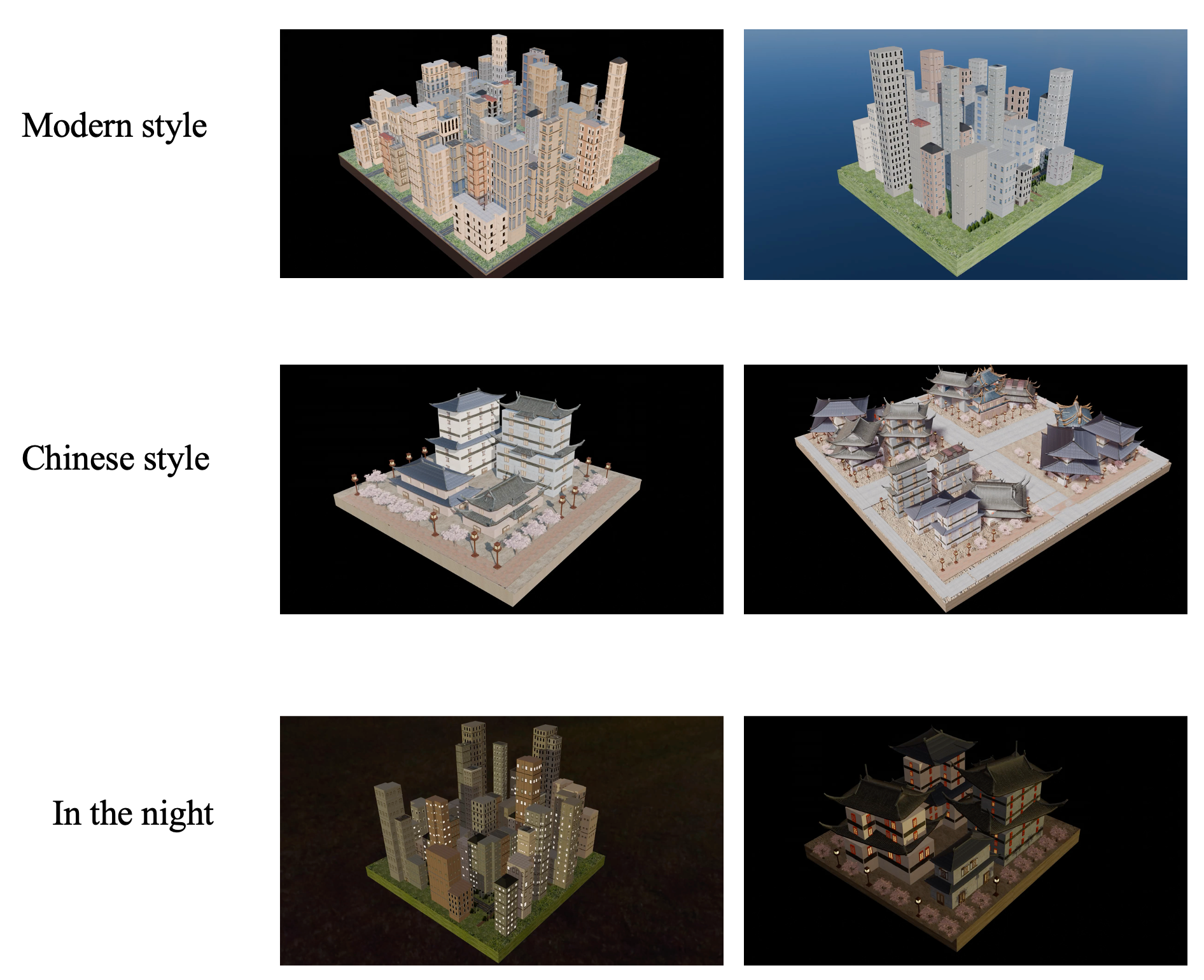}

 \caption{Generated City Results.}
 \label{scene_result}

\end{figure*}

\begin{figure*}[h]
    \centering
    \includegraphics[width=0.9\textwidth]{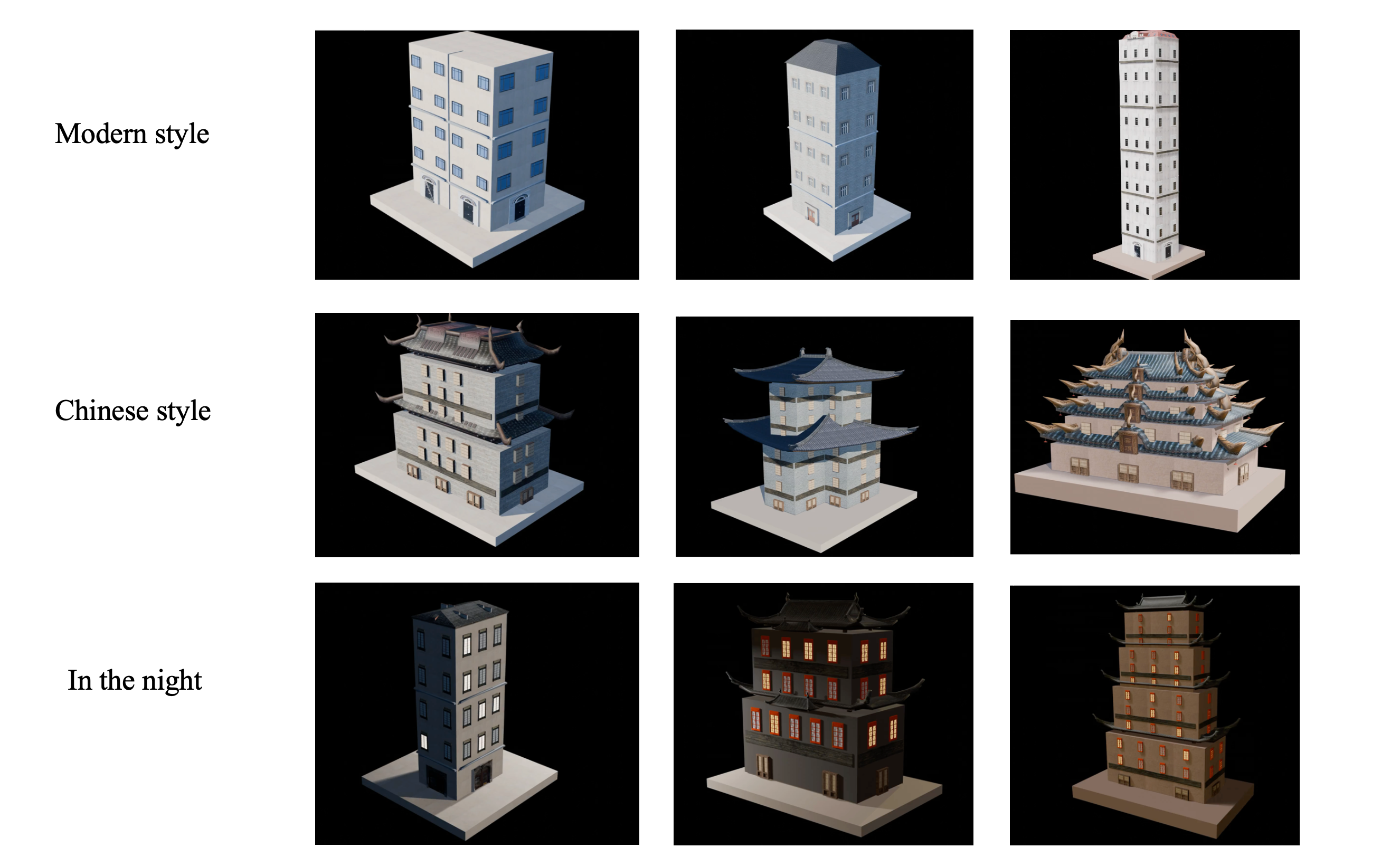}

 \caption{Generated Buildings Results.}
 \label{building_result}

\end{figure*}

\begin{figure*}[t]
   \centering
\includegraphics[width=0.8\textwidth]{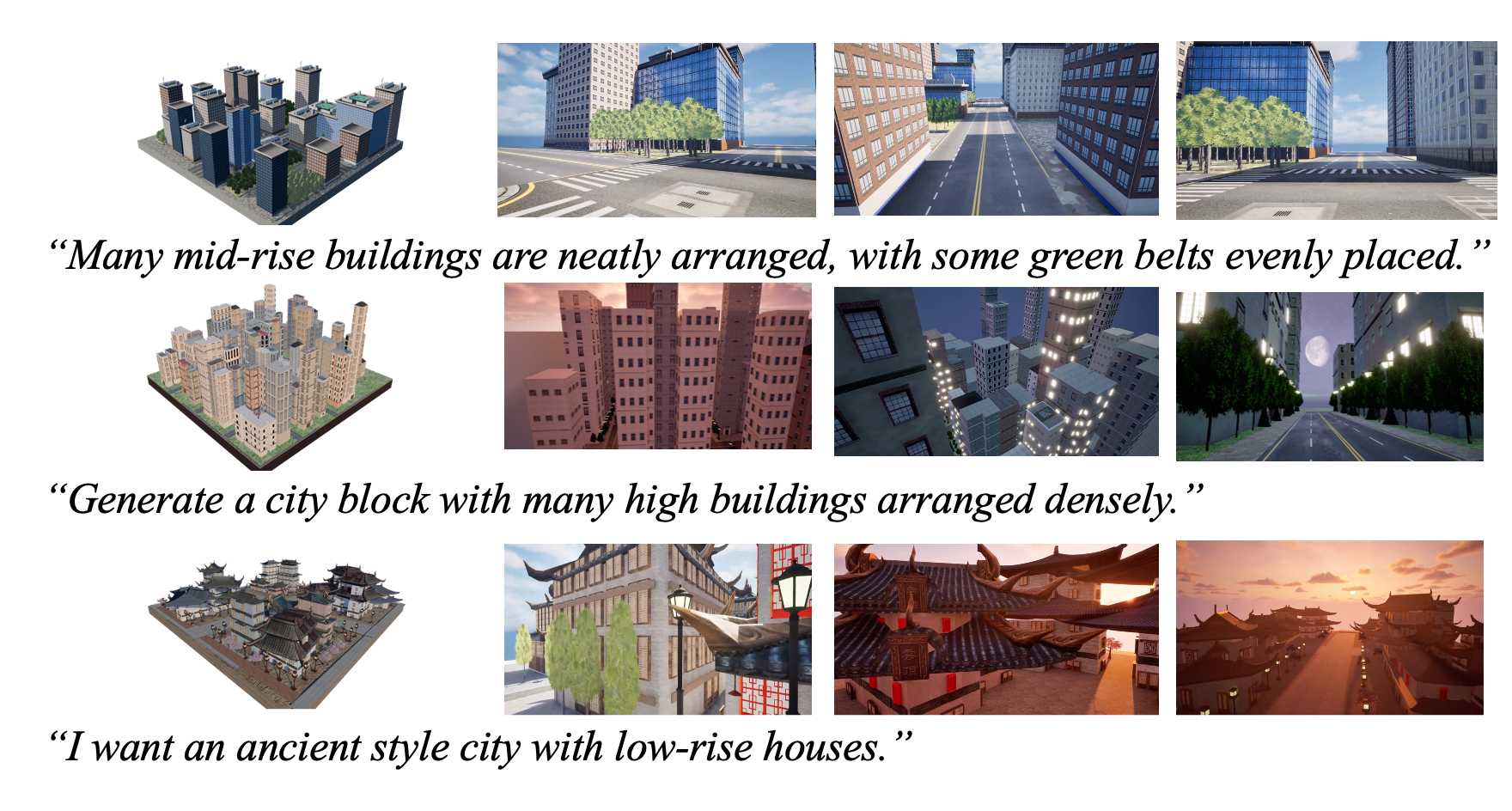}
 \caption{
Visual Results Generated by CityGenAgent. Results are shown across diverse conditions, including daytime, dusk, nighttime, and ancient style.}
 \label{visual_result}

\end{figure*}

\end{document}